\documentclass[letterpaper]{article} 
\usepackage{aaai25}  
\usepackage{times}  
\usepackage{helvet}  
\usepackage{courier}  
\usepackage[hyphens]{url}  
\usepackage{graphicx} 
\usepackage{natbib}  
\usepackage{caption} 
\usepackage{algorithm}
\usepackage{algorithmic}
\usepackage{booktabs} 
\usepackage{multirow} 
\usepackage{amsmath} 
\usepackage{amssymb}

\usepackage{newfloat}
\usepackage{listings}
\DeclareCaptionStyle{ruled}{labelfont=normalfont,labelsep=colon,strut=off} 
\floatstyle{ruled}
\newfloat{listing}{tb}{lst}{}
\floatname{listing}{Listing}
\nocopyright 
\title{HomoMatcher: Dense Feature Matching Results with 

Semi-Dense Efficiency by Homography Estimation}
\author{
    Xiaolong Wang\textsuperscript{\rm 1}\equalcontrib \thanks{Work done during internship at Ant Group.} \quad
    Lei Yu \textsuperscript{\rm 2}\equalcontrib \quad
    Yingying Zhang \textsuperscript{\rm 2}\quad
    Jiangwei Lao \textsuperscript{\rm 2}\quad
    Lixiang Ru \textsuperscript{\rm 2}\quad \\
    Liheng Zhong \textsuperscript{\rm 2}\quad
    Jingdong Chen \textsuperscript{\rm 2}\quad
    Yu Zhang \textsuperscript{\rm 1}\thanks{Corresponding Author.} \quad
    Ming Yang \textsuperscript{\rm 2}\footnotemark[3]
}
\affiliations{
    \textsuperscript{\rm 1}Zhejiang University  \quad
    \textsuperscript{\rm 2} Ant Group\\
    \{xlking,zhangyu80\}@zju.edu.cn, leiy@whu.edu.cn, m-yang4@u.northwestern.edu
}

\usepackage{bibentry}

\begin{document}

\maketitle

\begin{abstract}
Feature matching between image pairs is a fundamental problem in computer vision that drives many applications, such as SLAM. 
Recently, semi-dense matching approaches have achieved substantial performance enhancements and established a widely-accepted coarse-to-fine paradigm. 
However, the majority of existing methods focus on improving coarse feature representation rather than the fine-matching module.
Prior fine-matching techniques, which rely on point-to-patch matching probability expectation or direct regression, often lack precision and do not guarantee the continuity of feature points across sequential images.
To address this limitation, this paper concentrates on enhancing the fine-matching module in the semi-dense matching framework.
We employ a lightweight and efficient homography estimation network to generate the perspective mapping between patches obtained from coarse matching. 
This patch-to-patch approach achieves the overall alignment of two patches, resulting in a higher sub-pixel accuracy by incorporating additional constraints.
By leveraging the homography estimation between patches, we can achieve a dense matching result with low computational cost. 
Extensive experiments demonstrate that our method achieves higher accuracy compared to previous semi-dense matchers.
Meanwhile, our dense matching results exhibit similar end-point-error accuracy compared to previous dense matchers while maintaining semi-dense efficiency.
\end{abstract}
\section{Introduction}

Feature matching is a fundamental computer vision task that estimates pairs of pixels corresponding to the same 3D point from two images. This task is crucial for many downstream applications, such as Structure from Motion (SfM) \cite{schonberger2016structure, he2024detector}, Simultaneous Localization and Mapping (SLAM) \cite{mur2015orb, mur2017orb}, visual localization \cite{sarlin2019coarse, wang2024fine}, image stitching \cite{zaragoza2013projective}, etc.

Early approaches predominantly relied on feature detectors, which involved identifying salient points in a pair of images, crafting descriptors for these points, and subsequently accomplishing feature matching.
The focus during this period was on creating more efficient feature detectors, leading to the development of methods like SIFT \cite{lowe2004distinctive}, ORB \cite{rublee2011orb}, and other learning-based techniques \cite{detone2018superpoint}. However, the dependence on detectors significantly reduce robustness, resulting in failures in scenarios with textureless regions or large viewpint changes.

\begin{figure}[t]
\centering
\includegraphics[width=0.9\columnwidth]{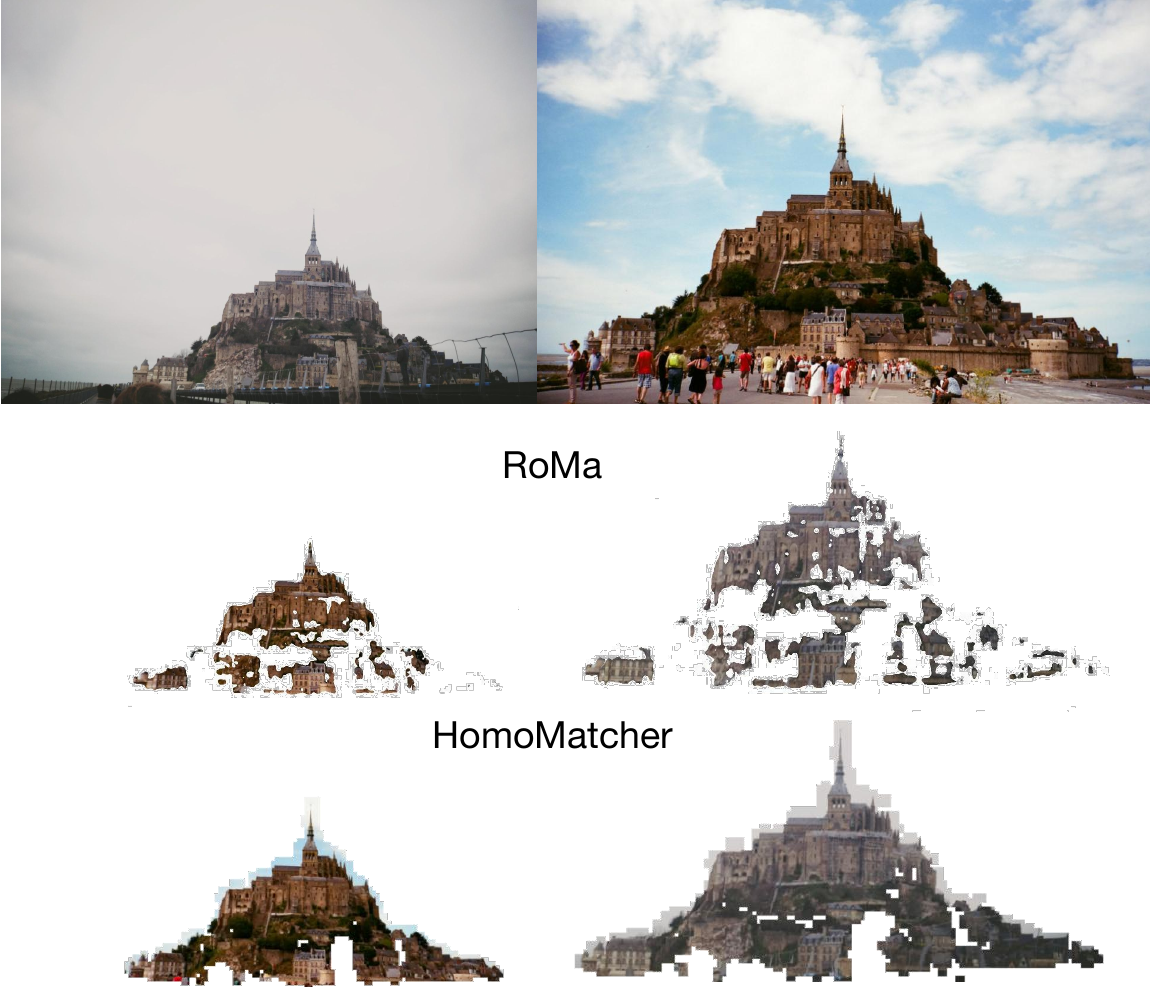} 
\caption{An visualization of dense matching results from our proposed HomoMatcher and dense matching method RoMa \cite{edstedt2024roma}. 
The HomoMatcher operates within a semi-dense framework, maintaining efficiency and enabling the flexible expansion of dense mappings from semi-dense results. 
Middle row is RoMa's results, which show warps with certainty values above a threshold of 0.02. 
Bottom row presents our results, demonstrating our method capability for dense matching refinement.}
\label{fig_1}
\end{figure}

Recently, LoFTR \cite{sun2021loftr} introduces a detector-free method based on a coarse-to-fine paradigm.
It leverages the context aggregation and positional encoding capabilities of Transformer \cite{vaswani2017attention} to generate discriminative coarse features, making it more adept at handling textureless scene. 
Mutual nearest neighbor strategies are employed to obtain coarse matches, which are then used to extract corresponding patch pairs from fine-level feature maps with high-resolution for further refinement. 
Fine-matching is performed based on the correlation and expectation calculated from the source patch center point and the target patch.
ASpanFormer \cite{chen2022aspanformer} processes an uncertainty-driven scheme to adaptively adjust local attention span, improving model performance through stronger feature representation.
Nevertheless, the fine-matching still relies on point-to-patch refinement.
This method of calculating expectation using point-to-patch matching can be influenced by irrelevant regions, leading to spatial variance that may affect fine-grained accuracy \cite{wang2024efficient}.

Several methods have also refined fine-level matching. 
Efficient LoFTR \cite{wang2024efficient} employs a two-stage refinement strategy to reduce the size of the corresponding patches, but it still relies on computing point-to-patch correlation expectations. 
\cite{chen2024affine} uses both patches for fine-matching, but directly regresses the offset of the source patch center without leveraging the geometric relationships between the two patches.
As a result, it still only achieves semi-dense matching with a single point per patch.

To address the aforementioned issues and considering the perspective transformation relationship among matched patches \cite{zaragoza2013projective}, we propose a lightweight yet effective homography estimation network to determine the fine-grained mapping between matched patch pairs. 
Our approach aligns patches by focusing on highly correlated regions, leveraging richer constraints to minimize the influence of irrelevant areas and achieve more accurate results.

With the obtained homography estimation, sparse or dense matching between the two patches can be performed freely and rapidly. 
Prior to this, detector-free methods like LoFTR encountered challenges in maintaining consistency of keypoints throughout sequential image matching in SLAM or SfM applications. Specifically, when an image is matched as a target at one moment and later served as a source, the resulting keypoints could be inconsistent, thereby impacting the Bundle Adjustment (BA) process during SLAM back-end optimization which needs a set of 2D
keypoint locations in multi-view images corresponding to the same 3D point \cite{peng2022rwt}. Our method can obtain match results from any position within the patches, ensuring continuity of keypoints during sequential matching.

Compared to dense matching methods \cite{edstedt2023dkm, edstedt2024roma}, our model maintains the efficiency of semi-dense approaches. 
The fine-matching module we proposed can be directly integrated into existing detector-free methods utilizing a coarse-to-fine framework. 
We conduct comprehensive experiments on the LoFTR and ASpanFormer models, demonstrating that our method significantly enhances model performance, even reaching state-of-the-art levels for semi-dense matching methods. 
Remarkably, our lightweight version also boosts the original model performances while maintaining faster processing speeds.

We also calculated the end-point error, a deterministic metric commonly used in the dense method \cite{edstedt2024roma}, to explicitly evaluate the model's performance in fine-grained matching.
The experimental results indicate that our method significantly outperforms other semi-dense approaches and achieves similar results to dense methods.

In summary, our main contributions are as follows:

\begin{itemize}
    \item We introduce a novel fine-matching module based on homography estimation, which suppresses spatial variance caused by irrelevant regions during refinement through patch-to-patch global alignment, achieving more accurate sub-pixel level matches.
    \item By leveraging homography estimation between patches, our method provides matching results for any point within the patch, ensuring keypoint repeatability. Additionally, it allows for densification of matches with the efficiency of semi-dense methods.
    \item The proposed method can be directly integrated into existing semi-dense approaches, and experiments demonstrate that replacing their fine-matching modules with our method significantly improves matching accuracy.
\end{itemize}
\section{Related Work}

\paragraph{Semi-Dense Image Matching.}
The semi-dense matching methods perform global matching solely at the downsampled coarse level, failing to deliver dense matching results at the original image resolution.
Approaches like \cite{rocco2018neighbourhood, rocco2020efficient, li2020dual} obtain image correspondences from 4D correlation volumes, but the limited receptive field of CNNs often leads to lower accuracy compared to sparse methods.
LoFTR \cite{sun2021loftr} first utilizes Transformer modules to semi-dense matching, enhancing performance through context-awareness and positional encoding. 
It employs a coarse-to-fine paradigm, initially enhancing features at the downsampled coarse level. 
These enhanced features are then used to obtain coarse matches through a mutual nearest neighbor mechanism. 
The matching is subsequently refined on high-resolution fine-level features using point-to-patch correlation.
Subsequent works have improved matching accuracy by enhancing coarse-level features further. 
Methods such as \cite{chen2022aspanformer, chen2024affine} use optical flow estimates to focus attention on relevant regions, enhancing the discriminability of coarse features. \cite{yu2023adaptive} introduces a spot-guided aggregation module to minimize the impact of irrelevant areas during feature aggregation, though it still uses LoFTR refinement approach. \cite{wang2024efficient} notes that LoFTR's refinement is affected by irrelevant regions leading to positional variance and proposes a two-stage approach: further matching fine-level patch pairs using MNN and refining within a smaller region using refinement-by-expectation. 
Motivated by similar goals, we globally align patches and focus on highly correlated areas to reduce positional variance, significantly improving matching accuracy. 
Additionally, unlike previous methods that require Transformer-based enhancements or feature fusion for fine-level features before fine-matching, our approach directly utilizes fine-level features extracted by the backbone.
And our method achieves dense matching by obtaining all correspondences within the patches, as shown in Figure \ref{fig_1}.

\begin{figure*}[t]
\centering
\includegraphics[width=1.9\columnwidth]{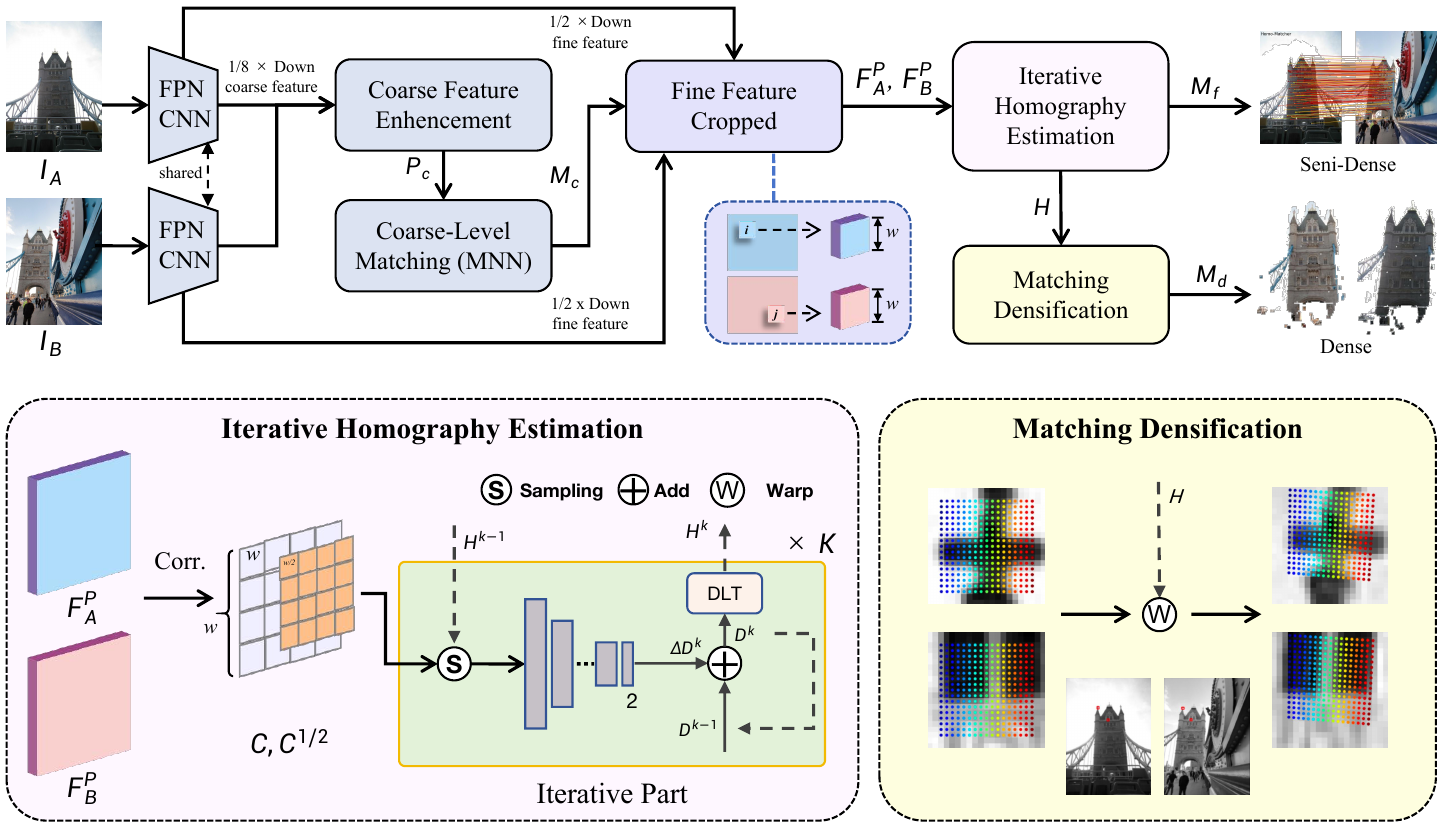}
\caption{An illustration of the proposed method.We use a CNN backbone to extract coarse-level and fine-level features from the given images $I_A$ and $I_B$. Initially, we enhance the coarse features and then obtain the coarse matching result $\mathcal{M}_c$ using the MNN criterion. For each match $(i,j) \in \mathcal{M}_c$, we extract patch pairs of size $w \times w$ from the fine-level features centered at the upsampled position, resulting in $F_A^P$ and $F_B^P$. We estimate the homography $\mathbf{H}$ between matched patches iteratively to refine the subpixel-level matches, yielding the refined matches $\mathcal{M}_f$. We can also obtain the corresponding dense matches $\mathcal{M}_d$.}
\label{fig:overview}
\end{figure*}

\paragraph{Dense Image Matching.}
Dense matching methods aim to obtain all pixel correspondences between images. DKM \cite{edstedt2023dkm} models dense matching as a probabilistic problem and estimates a dense certainty map to filter matching results. It follows a coarse-to-fine paradigm, refining results through a multi-scale refinement module that continuously upsamples previous matches and inputs them along with fine-level features into a CNN for regression. \cite{edstedt2024roma} builds on DKM, using a more powerful feature encoder and replacing the original CNN-based decoder with a transformer-based decoder during coarse matching. \cite{zhu2023pmatch} replaces DKM probabilistic regression with a correlation-based matching process and applies a similar refinement strategy. Although these dense methods offer higher accuracy, they are computationally expensive and often slow for many practical tasks. 
Our proposed method, based on a semi-dense framework, achieves dense matching results with comparable pixel-level accuracy, while being significantly faster.

\paragraph{Homography Estimation}
Traditional homography estimation methods involve detecting and matching feature points, outlier removal, and computing the homography matrix using Direct Linear Transformation (DLT) \cite{hartley2003multiple}. 
\cite{detone2016deep} pioneered deep learning-based homography estimation methods. 
\cite{nguyen2018unsupervised} introduced an unsupervised homography estimation approach by calculating photometric errors. 
\cite{zhao2021deep} incorporated the Lucas-Kanade (LK) algorithm \cite{lucas1981iterative} for iterative homography estimation. 
IHN \cite{IHN} further improves homography estimation performance using global motion aggregation and correlation calculations.
HomoGAN \cite{hong2022unsupervised} introduces an unsupervised homography estimation network based on Transformers and GANs, which improves performance but significantly increases computational cost.
\section{Method}

\subsection{Preliminary} 

As illustrated in Figure \ref{fig:overview}, our approach adopts the coarse-to-fine paradigm pioneered by LoFTR \cite{sun2021loftr}. Given a pair of images, $I_A$ and $I_B$, our network generates coarse matches at a downsampled resolution, which are then refined using homography estimation. 
Initially, both images are processed through a ResNet backbone with a Feature Pyramid Network (FPN) for multi-level feature extraction. 
The coarse-level features are extracted at 1/8 of the original resolution, while the fine-level features are at 1/2.

These coarse features undergo feature enhancement through iterative self/cross attention modules after positional encoding, implemented via a transformer. Some recent methods employ adaptive attention areas \cite{chen2022aspanformer} or Deformable Attention \cite{chen2024affine} to further enhance feature representation. 
Upon obtaining discriminative coarse features, a score matrix is derived from the inner product of features, and a preliminary match probability matrix $\mathcal{P}_{c}$ is obtained using a dual-softmax operator. 
Next, coarse matching results ${\mathcal{M}_{c}}$ are determined using Mutual-nearest-neighbor (MNN):

\begin{equation}
\mathcal{M}_{c}=\left\{
(i, j) \mid \forall(i, j) \in \operatorname{MNN} ( \mathcal{P}_{c} ), \mathcal{P}_{c}(i, j) \geq \theta_{c}
\right\}
\end{equation}

where $i, j$ represent positions on the 1/8 downsampled images of $I_A$ and $I_B$, respectively, and $\theta_{c}$ is the probability threshold for coarse matching. 
To achieve sub-pixel accurate matching, feature patches are cropped from the fine-level features centered around ${\mathcal{M}_{c}}$ for refinement. 
Previous methods would choose a reference point in the source patch during the fine-matching stage, followed by feature correlation and exception. 
However, this method can be influenced by irrelevant regions, contributing to spatial variance.

As demonstrated in \cite{zaragoza2013asap}, small patches between images can be successfully aligned using homography estimation.
We propose a method to align patches using homography estimation, focusing only on regions with high correlation and ignoring those with low relevance. 
This alignment of highly correlated regions between patches results in a mapping matrix, leading to more precise and robust outcomes. 
Details are provided in the following sections.

\subsection{Homography Estimation for Fine Matching}

We extract patch pairs of size $w \times w$ from the fine-level features centered around the coarse match results $(i, j)$, and view the number of matches to the batch dimension for processing. 
Given $N$ matches from coarse matching, the resulting patch features are $F^P_A, F^P_B \in \mathbb{R}^{N \times D \times w \times w}$, where we set $w$ as a fixed value.

Inspired by direct SLAM methods \cite{engel2014lsd}, we aim to optimize a homography matrix to minimize the differences between corresponding regions after mapping the patches. 
Unlike traditional SLAM methods that optimize based on photometric error, our approach uses correlation to refine and update the homography. 
We sample correlations using the latest estimated homography matrix and input them as feature values into the network, effectively reducing the influence of irrelevant areas due to their low correlation values.

\paragraph{Correlation Computation}  

\begin{figure}[th]
\centering
\includegraphics[width=0.9\columnwidth]{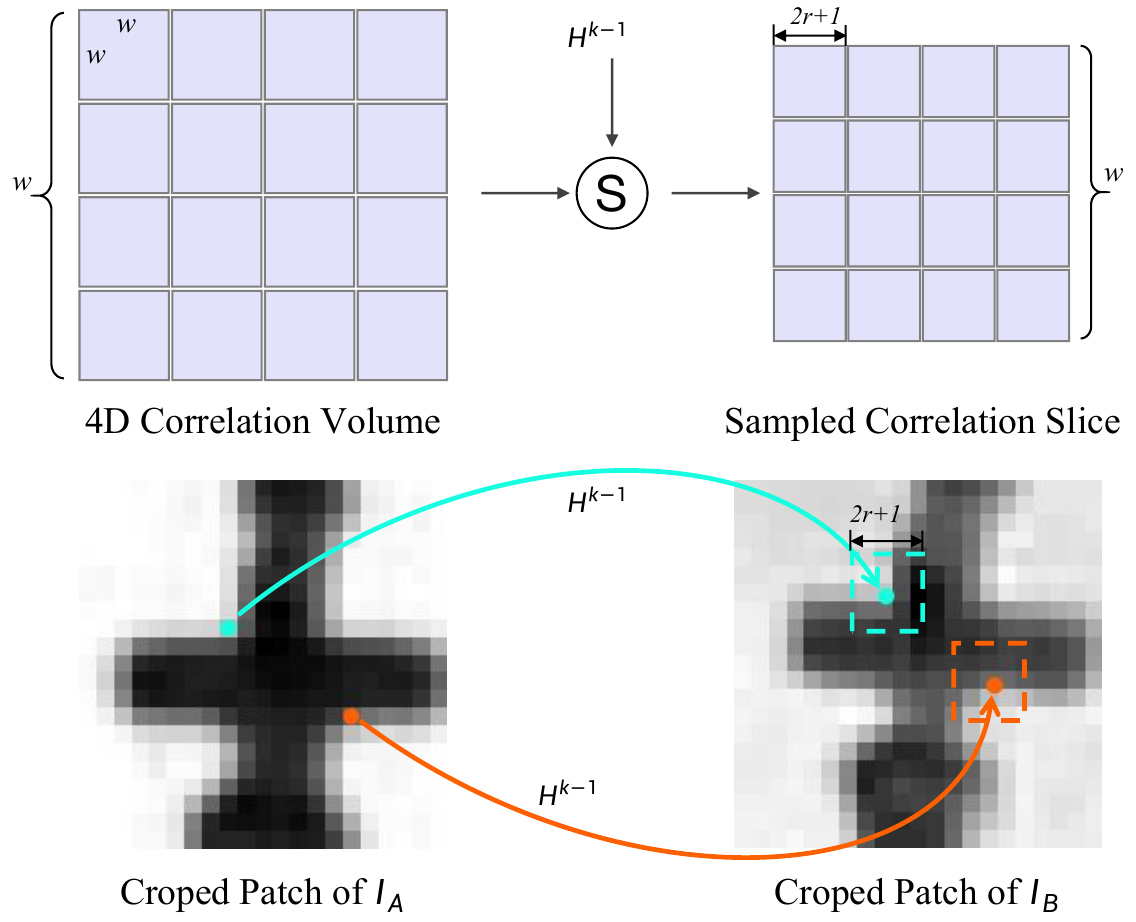} 
\caption{Visualization of sampling a 4D correlation volume using homography estimation $H^{k-1}$.
The top row illustrates the process of sampling a 4D correlation volume, which has dimensions $w \times w \times w \times w$, into a $w \times w \times (2r+1) \times (2r +1) $ 4D correlation slice.
The bottom row demonstrates how each pixel location is sampled from the correlation patch using a $(2r+1) \times (2r +1)$ window based on pixel mapping results.
}
\label{fig:corr}
\end{figure}

Based the patch features, we create the correlation volume $\textbf{C} \in \mathbb{R}^{w \times w \times w \times w}$ by computing the dot product between all feature vector pairs:

\begin{equation}
    C_{i j k l}=\operatorname{ReLU}(F^P_A(i,j)^\mathrm{T} F^P_B(k,l)).
    \label{equ:correlation}
\end{equation}

To extend the receptive field , we apply average pooling to $\textbf{C}$ over the last 2 dimensions with a stride of 2, producing an additional correlation volume $\textbf{C}^{\frac{1}{2}}  \in \mathbb{R}^{w \times w \times w/2 \times w/2}$.

\paragraph{Iterative Homography Estimation}

We iteratively estimate the homography $\textbf{H}$, initially set as the identity matrix. 
A unit coordinate grid $\textbf{X} \in \mathbb{R}^{2 \times w \times w}$ is generated on the reference patch $F^P_A$ and projected onto $\textbf{X}' \in \mathbb{R}^{2 \times w \times w}$ on $F^P_B$ using the current estimate of $\textbf{H}$.
For each coordinate position $x = (u, v)$ in $\textbf{X}$ and $x' = (u', v')$ in $\textbf{X}'$, the mapping is performed using Equation \ref{equ:homoproject}. 
We then sample the 4D correlation volume $\textbf{C}$ with $\textbf{X}'$ using a local square grid of fixed search radius $r$, resulting in correlation slices $\textbf{S}^k$ of size $w \times w \times (2r+1) \times (2r+1)$.  
For the 1/2 down-sampled $\textbf{C}^{\frac{1}{2}}$, we scale $\textbf{X}'$ by a factor of 0.5 and apply bilinear interpolation to sample and obtain $\textbf{S}^{\frac{1}{2}, k}$.
The sampling process is illustrated in Figure \ref{fig:corr}.

\begin{equation}
    \left[\begin{array}{c}
        u^{\prime k} \\
        v^{\prime k} \\
        1
        \end{array}\right] \sim\left[\begin{array}{ccc}
        \mathbf{H}_{11}^{k} & \mathbf{H}_{12}^{k} & \mathbf{H}_{13}^{k} \\
        \mathbf{H}_{21}^{k} & \mathbf{H}_{22}^{k} & \mathbf{H}_{23}^{k} \\
        \mathbf{H}_{31}^{k} & \mathbf{H}_{32}^{k} & 1
        \end{array}\right]\left[\begin{array}{l}
        u \\
        v \\
        1
    \end{array}\right]
    \label{equ:homoproject}
\end{equation}

The homography matrix is parameterized by displacement vectors at the four corners of the patch, represented as the displacement cube $\textbf{D}$. 
We estimate the homography residual using a CNN-based decoder. 
At the $k$-th iteration, the decoder takes as input the concatenation of the correlation slices $\textbf{S}^k$, $\textbf{S}^{\frac{1}{2}}$, and the coordinates $\textbf{X}$ and $\textbf{X}'^k$. 
The decoder consists of several convolutional-based units, each reducing the spatial resolution by a factor of 2 until it reaches a $2 \times 2$ resolution. 
A final $1 \times 1$ convolutional layer projects the feature map into a $2 \times 2 \times 2$ cube $\Delta \textbf{D}^k$, representing the estimated residual displacement vectors for the four corners. 
The displacement cube $\textbf{D}^k$ is updated by adding $\Delta \textbf{D}^k$ to $\textbf{D}^{k-1}$. 
The updated $\textbf{H}^k$ is derived from $\textbf{D}^k$ using direct linear transformation \cite{abdel2015direct} and is then used to project $\textbf{X}$ in the next iteration.

\subsection{Matching Densification}

\begin{figure}[t]
\centering
\includegraphics[width=0.9\columnwidth]{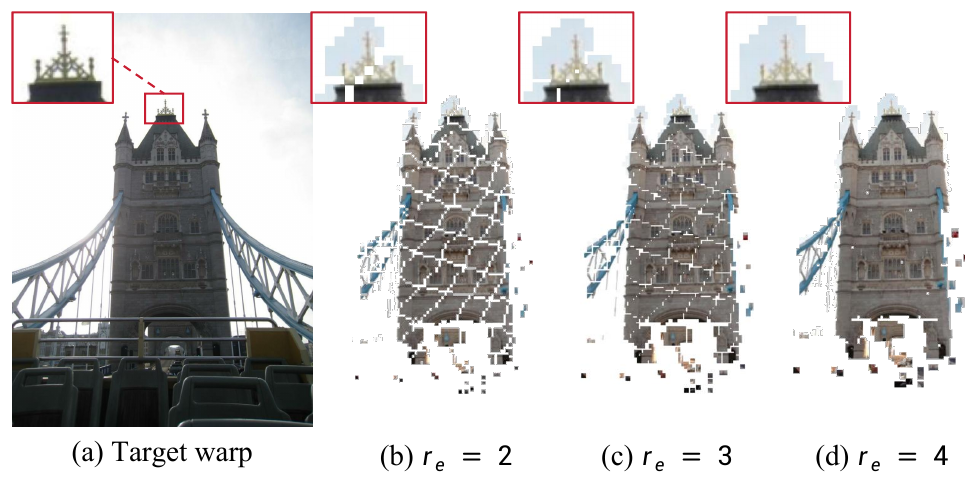} 
\caption{Visualization of the impact of different expansion radius ($r_e$) on match densification. 
From left to right, the images show the warp target and the warp results of dense matching obtained with expansion radius of $r_e = 2, 3, 4$. The zoomed details of the spire further illustrate the reliability of our model's densification.}
\label{fig:dense_re}
\end{figure}

After $K$ iterations, the homography transformation matrix $\mathbf{H}$ for each pair of patches is obtained. 
The center of the source patch can then be mapped using Equation \ref{equ:homoproject} to achieve fine-level matching results $\mathcal{M}_f$ consistent with the previous fine-matching module. 
Additionally, for sequential images, the matching result of $I_A$ at the previous time instance can also be mapped onto the corresponding patch, ensuring keypoint repeatability.

Based on the obtained patch-level homography transformation, we can also easily achieve dense matching. Since the fine-level is at a resolution of $1/2$, we can generate an expanded grid $G_e$ centered at each patch with a step of 0.5 to obtain a dense matching result that can be up-sampled to the original size of the image. For example, if the expansion radius $r_e = 0.5$ on the fine-level patch, then:

\begin{equation}
    G_e = \begin{bmatrix}
 (-0.5, -0.5) & (0, -0.5) & (0.5, -0.5) \\
 (-0.5, 0) & (0, 0) & (0.5, 0) \\
 (-0.5, 0.5) & (0, 0.5) & (0.5, 0.5)
\end{bmatrix}.
\end{equation}

Let $\mathbf{X} = G_e + \lfloor \frac{w}{2} \rfloor$ denote the coordinates on $F_P^A$, we can use $\mathbf{H}$ to obtain dense matching results $\mathcal{M}_d$. 

Matches $\mathcal{M}_{c}$ are obtained at a $1/8$ resolution. 
Ideally, $r_e = 2$ ensures dense matching results. 
However, since coarse matching $\mathcal{M}_{c}$ uses the Mutual Nearest Neighbor (MNN) method, many-to-one matching results are suppressed to at most one, so $r_e$ may need to be increased to achieve dense matching. 
The effects of different $r_e$ values on densification are shown in Figure \ref{fig:dense_re}.
In cases where the densification results of different patches overlap, we select the mapping from the patch whose center is closest to the overlapping point as the final matching result.

\subsection{Supervision}

Our proposed fine-matching module can be seamlessly integrated into existing coarse-to-fine semi-dense matching models.
These models typically employ a hybrid loss function that incorporates both coarse and fine matching results for training supervision. 
During training, we retain all original losses except for the fine-level loss, defining the overall loss as $\mathcal{L} = \mathcal{L}_{c} + \mathcal{L}_{f}$. 
$\mathcal{L}_{c}$ includes losses associated with the coarse matching stage, such as the log-likelihood loss computed using the coarse ground truth ${\mathcal{M}_{gt}}$, and in some cases, like in AspanFormer \cite{chen2022aspanformer}, an additional optical flow loss derived from the optical flow obtained during the coarse matching phase.

The fine-level loss is calculated directly by computing the L2 loss between the refined matching results and the ground truth. All coordinates are normalized according to the size and center points of the respective patches to maintain a consistent scale with the coarse-level loss. Here, we use the densified refined matching for loss calculation, with each patch being supervised by $(2 \times r_e / 0.5 + 1)^2$ point pairs. Homography estimation can typically be computed with four point pairs, thus the densified $\mathcal{L}_{f}$ ensures effective supervision for the homography estimation.

Furthermore, previous methods use coarse matches obtained through Mutual Nearest Neighbor (MNN) for fine-level supervision, which suppressed one-to-many patch pair results. 
This approach is beneficial for coarse matching supervision as it eliminates matching ambiguities. 
However, it limits the diversity of samples available for homography estimation.
To address this, we have additionally incorporated the suppressed yet correct matches, which are excluded by MNN, into the supervision of the fine-matching module.

\begin{table*}[t]
\centering
\begin{tabular}{@{}cccccccc@{}}
\toprule
\multirow{2}{*}{Category}   & \multirow{2}{*}{Method}   & \multicolumn{3}{c}{MegaDepth}                                             & \multicolumn{3}{c}{ScanNet}                          \\ \cmidrule(l){3-8} 
                            &                           & AUC@5↑               & AUC@10 ↑      & \multicolumn{1}{c|}{AUC@20 ↑}      & AUC@5↑               & AUC@10 ↑      & AUC@20 ↑      \\ \midrule
Sparse                      & SP+SG                     & 49.7                 & 67.1          & \multicolumn{1}{c|}{80.6}          & 16.2                 & 32.8          & 49.7          \\ \midrule
\multirow{7}{*}{Semi-Dense} & LoFTR                     & 52.8                 & 69.2          & \multicolumn{1}{c|}{81.2}          & 16.9                 & 33.6          & 50.6          \\
                            & ASpanFormer               & 55.3                 & 71.5          & \multicolumn{1}{c|}{83.1}          & 19.6                 & 37.7          & 54.4          \\
                            & LoFTR (E)                 & 56.4                 & 72.2          & \multicolumn{1}{c|}{83.5}          & 19.2                 & 37.0          & 53.6          \\
                            & Affine                    & 57.3                 & 72.8          & \multicolumn{1}{c|}{84.0}          & 22.0                 & \textbf{40.9} & \textbf{58.0} \\
                            & LoFTR\_Homo${}^{*}$       & 55.1 (↑2.3)          & 71.8          & \multicolumn{1}{c|}{83.4}          & 18.4 (↑1.5)          & 35.4          & 51.8          \\
                            & ASpan\_Homo${}^{*}$       & 57.1 (↑1.8)          & 73.0          & \multicolumn{1}{c|}{84.1}          & 22.0 (↑2.4)          & 40.5          & 57.2          \\
                            & ASpan\_Homo${}^{\dagger}$ & \textbf{57.8 (↑2.5)} & \textbf{73.5} & \multicolumn{1}{c|}{\textbf{84.4}} & \textbf{22.1 (↑2.5)} & \textbf{40.9} & 57.5          \\ \midrule
\multirow{2}{*}{Dense}      & DKM                       & 60.4                 & 74.9          & \multicolumn{1}{c|}{85.1}          & 26.6                 & 47.2          & 64.2          \\
                            & RoMA                      & 62.6                 & 76.7          & \multicolumn{1}{c|}{86.3}          & 28.9                 & 50.4          & 68.3          \\ \bottomrule
\end{tabular}
\caption{Results of two-view pose estimation on the MegaDepth \cite{li2018megadepth} and ScanNet \cite{dai2017scannet} datasets.
Both datasets are evaluated using the same model trained on MegaDepth. 
$*$ indicates that our fine-matching module is inserted, and $\dagger$ denotes the heavy version.
Improvements compared to the original method are shown in parentheses.
}
\label{tab:auc_table}
\end{table*}

\section{Experiments}

\subsection{Implementation Details}
Following \cite{wang2024efficient}, our model is trained on the MegaDepth dataset \cite{li2018megadepth}, which consists of outdoor scenes.
All subsequent experiments are conducted on this trained model to evaluate its generalization capabilities. 
We integrate the homography estimation fine-matching module into LoFTR \cite{sun2021loftr} and ASpanFormer \cite{chen2022aspanformer} for separate training sessions, naming the models LoFTR\_Homo and ASpan\_Homo, respectively.
The threshold for obtaining coarse matches is set at $\theta_c = 0.2$. 
The size of patches cropped from the fine-level features is $w=9$. 
The correlation search radius during iterative homography estimation is $r=1$, with $K=3$ iterations. 
The densification radius used during loss calculation is $r_e=2$. 
The model training utilizes the Adam optimizer with a learning rate of $1 \times 10^{-3}$ and a batch size of 8 for 30 epochs on 8 V-100 GPUs.
Additionally, we train a heavy version with parameters $w=17$, $r=3$, and $K=3$ to investigate the full capabilities of the model, indicated by $\dagger$.

\subsection{Two-view Pose Estimation}
\paragraph{Datasets.}
We employ the MegaDepth \cite{li2018megadepth} and ScanNet \cite{dai2017scannet} datasets to validate our model's matching capability for relative pose estimation in both outdoor and indoor settings. 
MegaDepth consists of 1 M image pairs from 196 3D scenes, providing ground truth relative poses and depth maps calculated using COLMAP. We follow the training and validation splits used in previous methods \cite{sun2021loftr}, with the validation set comprising 1500 randomly selected image pairs from scenes 0015 and 0022. 
Images are resized with the longest edge to 832 and 1152 for training and validation, respectively.

ScanNet is a large-scale indoor dataset containing 1613 image sequences, presenting challenges due to textureless surfaces and varying viewpoints. We evaluate using the same 1500 test image pairs as \cite{sarlin2020superglue}, with all test images resized to $480 \times 640$.

\begin{figure*}[t]
\centering
\includegraphics[width=1.9\columnwidth]{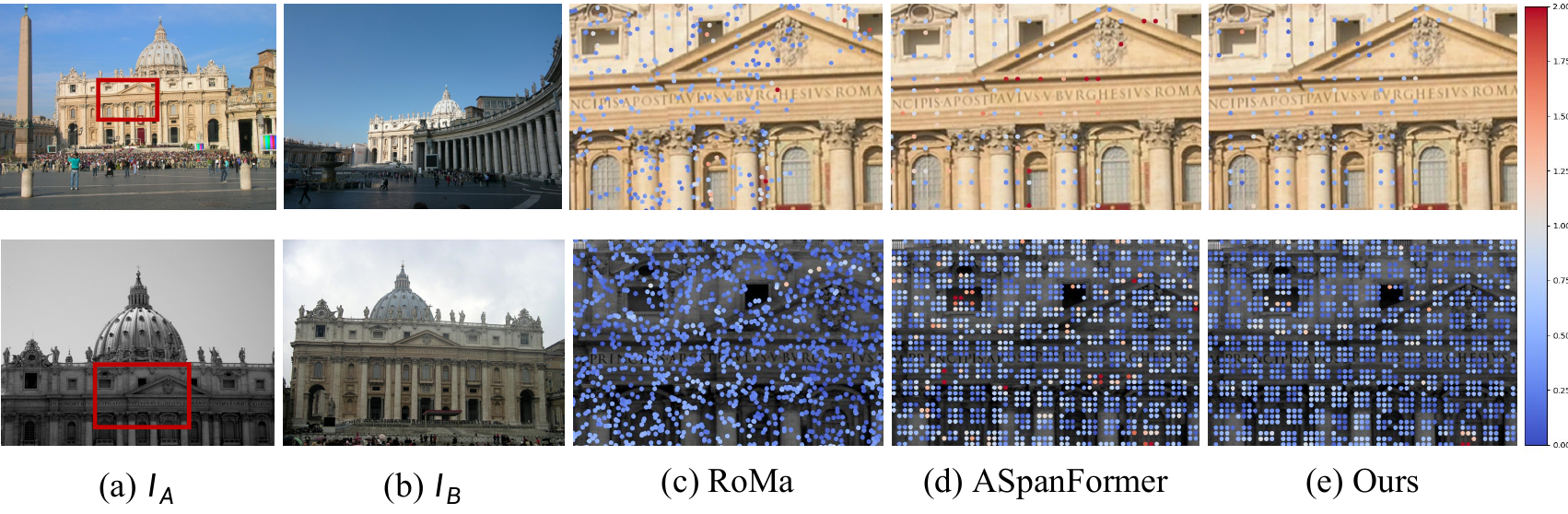} %
\caption{Visualization of end-point-error(EPE). 
The color gradient from blue to red represents increasing EPE. 
}
\label{fig:vis_epe}
\end{figure*}

\paragraph{Comparative methods.}

We compared our method against three types of approaches:
1) Sparse methods based on the SuperPoint \cite{detone2018superpoint} keypoint detector and SuperGlue \cite{sarlin2020superglue} matcher,
2) Detector-free semi-dense matchers, including LoFTR \cite{sun2021loftr}, ASpanFormer \cite{chen2022aspanformer}, Efficient LoFTR \cite{wang2024efficient}, and Affine-based Matcher \cite{chen2024affine},
3) Dense matchers, including DKM \cite{edstedt2023dkm} and RoMa \cite{edstedt2024roma}.

\paragraph{Evaluation protocol}

As with previous methods, we use the matching results to compute the essential matrix and recover the relative poses. This computation is performed using OpenCV's RANSAC implementation with settings consistent with \cite{sun2021loftr}. We report the AUC of the pose error at thresholds of 5, 10, and 20 degrees.

Furthermore, we note that relative pose estimation introduces randomness, and the differences between methods significantly diminish when using advanced RANSAC techniques \cite{wang2024efficient}. 
Therefore, we additionally measure the end-point error (EPE) on testing split of MegaDepth dataset, reporting the percentage of correct keypoints at specific pixel thresholds, known as the percent correct keypoint (PCK) \cite{edstedt2024roma}.
To ensure a fair comparison across different methods, we use the pixel error between the sparse matching points used for calculating the relative pose and the ground-truth correspondences at the original image resolution. 
Consistent with \cite{edstedt2024roma}, we set the thresholds at 1px, 3px, and 5px.

\paragraph{Results.}

As for the relative pose metrics shown in Tab. \ref{tab:auc_table}, merely replacing the fine-matching module with our proposed homography estimation method has significantly improved the performance of both LoFTR \cite{sun2021loftr} and ASpanFormer \cite{chen2022aspanformer} on two datasets. 
On the MegaDepth dataset, the AUC@5 for ASpan\_Homo based LoFTR improved by 4.4\%, and for ASpan\_Homo, it increased by 3.3\%. 
Furthermore, the ASpan\_Homo can achieve performance comparable to the previous state-of-the-art semi-dense methods.

The PCK results in Tab. \ref{tab:epe_table} and the qualitative examples in Figure \ref{fig:vis_epe} demonstrate that our method significantly outperforms previous semi-dense matching methods in pixel-level accuracy. 
Dense methods, like RoMa \cite{edstedt2024roma} with its DINOv2 backbone, often rely on heavier and more complex architectures, leading to slower runtimes in practical applications. 
In contrast, our approach achieves comparable fine-grained matching accuracy with significantly lower computational cost. 
Furthermore, the heavy version surpasses DKM \cite{edstedt2023dkm} in accuracy. 
On a single V100 GPU with MegaDepth images resized to 1152 resolution, ASpan\_Homo runs in 442 ms, the heavy version in 697 ms, while RoMa, using the official code, takes 1527 ms.

\begin{table}[t]
\centering
\begin{tabular}{@{}lllll@{}}
\toprule
Method                    & @1px ↑        & @3px ↑        & @5px ↑        & Runtime↓ \\ \midrule
LoFTR                     & 50.6          & 86.8          & 92.8          & 350 ms   \\
ASpanFormer               & 54.3          & 90.1          & 95.6          & 412 ms   \\
LoFTR (E)                 & 54.4          & 88.6          & 93.3          & 178 ms   \\
LoFTR\_Homo${}^{*}$       & 57.5          & 88.9          & 93.5          & 379 ms   \\
ASpan\_Homo${}^{*}$       & 60.2          & 91.6          & 96.1          & 442 ms   \\
ASpan\_Homo${}^{\dagger}$ & \textbf{62.5} & \textbf{92.5} & \textbf{96.5} & 697 ms   \\ \midrule
DKM                       & 62.0          & 90.7          & 94.8          & 1175 ms  \\
RoMA                      & 63.7          & 92.0          & 96.0          & 1527 ms  \\ \bottomrule
\end{tabular}
\caption{
Results for end-point error (EPE) on MegaDepth. 
All methods are measured using the fine-level matching at the real image resolution, computing pixel offset. 
We report the percentage of correct keypoints (PCK) with an EPE less than 1 pixel, 3 pixels, and 5 pixels.}
\label{tab:epe_table}
\end{table}

\subsection{Homography Estimation}
\begin{table}[t]
\centering
\begin{tabular}{@{}llll@{}}
\toprule
Method         & @3px ↑ & @5px ↑ & @10px ↑ \\ \midrule
SuperGlue      & 53.9 & 68.3 & 81.7  \\
LoFTR          & 65.9 & 75.6 & 84.6  \\
ASpanFormer    & 67.4 & 76.9 & 85.6  \\
LoFTR (E) & 66.5 & 76.4 & 85.6  \\
ASpan\_Homo${}^{*}$    & 70.2 & 79.6 & 87.8  \\ \bottomrule
\end{tabular}
\caption{Homography Estimation results on Hpatches dataset \cite{balntas2017hpatches}.}
\label{tab:hpatch_table}
\end{table}

The HPatches dataset \cite{balntas2017hpatches} provides multiple sets of sequential images of the same scenes under varying viewpoints and illumination conditions, along with the corresponding ground truth homographies. 
Following the evaluation protocol of LoFTR \cite{sun2021loftr}, we resize the shorter side of the images to 480. 
We compute the mean error of corner points and report the Area Under Curve (AUC) at three pixel thresholds: 3px, 5px, and 10px, using the same RANSAC method employed by other approaches to performing homography estimation. 
The experimental results in Tab. \ref{tab:hpatch_table} show that our method significantly outperforms previous sparse and semi-dense methods.

\subsection{Ablation Study}

\paragraph{Fine-level Supervision.} 
Benefiting from our model's capability to generate dense matching results, we are able to compute a dense fine-level loss. 
We conducted ablation studies to compare semi-dense supervision, where each coarse-matched patch pair produces only one refinement matching point, against dense supervision.
Additionally, we performed an experiment to assess the impact of incorporating suppressed yet correct matches into the supervision.
The results in Tab. \ref{tab:loss_table} demonstrate that densified loss significantly improves the pixel accuracy of fine matching outcomes, and increasing sample diversity during the refinement training phase can further boost the model's performance.

\begin{table}[t]
\centering
\begin{tabular}{@{}lllll@{}}
\toprule
\multirow{2}{*}{Method} & \multicolumn{2}{l}{EPE PCK↑} & \multicolumn{2}{l}{Pose Est AUC} \\ \cmidrule(l){2-5} 
                        & @1px          & @3px         & @5              & @20            \\ \midrule
semi-dense fine loss      & 58.6          & 90.5         & 56.5            & 83.8           \\
+ dense fine loss       & 59.1          & 90.8         & 57.0            & 84.0           \\
+ suppressed matches    & 60.2          & 91.6         & 57.1            & 84.1           \\ \bottomrule
\end{tabular}
\caption{Loss ablation study on MegaDepth.}
\label{tab:loss_table}
\end{table}

\paragraph{Efficiency Evaluation}

Within the ASpanFormer \cite{chen2022aspanformer} framework, we conducted experiments on the original fine-matching module and our proposed homography-based fine-matching module with various hyperparameters ($w, r, K$), as shown in Tab. \ref{tab:param_table}. 
Since our improvements are specific to the fine-matching process, we only report the runtime for this module. 
It is evident that heavier model parameters lead to higher accuracy, but also result in increased runtime. 
Notably, compared to the original fine-matching module, the model with 5/1/1 parameters achieves higher matching accuracy more efficiently while also maintaining the capability for densification.
This demonstrates the potential of our proposed method to become the optimal choice for the fine-matching module within the semi-dense matching paradigm. 
Furthermore, our model demonstrates that increasing complexity can yield corresponding performance improvements, offering options for offline tasks that prioritize accuracy over latency.

\begin{table}[t]
\centering
\begin{tabular}{@{}llllll@{}}
\toprule
\multirow{2}{*}{$w/r/K$} & \multicolumn{2}{c}{EPE PCK↑}                        & \multicolumn{2}{c}{Pose Est AUC↑}                & \multicolumn{1}{c}{\multirow{2}{*}{Runtime ↓}} \\ \cmidrule(lr){2-5}
                         & \multicolumn{1}{c}{@1px} & \multicolumn{1}{c}{@3px} & \multicolumn{1}{c}{@5} & \multicolumn{1}{c}{@20} & \multicolumn{1}{c}{}                         \\ \midrule
orig                     & 54.3                     & 90.1                     & 55.3                   & 83.1                    & 29.4 ms                                      \\
5/1/1                    & 55.6                     & 90.6                     & 56.2                   & 83.3                    & \textbf{13.7 ms}                                      \\
9/1/1                    & 58.5                     & 91.5                     & 56.5                   & 83.9                    & 38.7 ms                                      \\
9/1/3                    & 60.2                     & 91.6                     & 57.1                   & 84.1                    & 58.7 ms                                      \\
9/3/1                    & 59.9                     & 91.9                     & 57.3                   & 84.0                    & 53.4 ms                                      \\
9/3/3                    & 60.3                     & 91.9                     & 57.1                   & 84.2                    & 101.4 ms                                     \\ \bottomrule
\end{tabular}
\caption{Hyper-parameter ablation study on MegaDepth.
We use one V100 GPU and 1152 resolution for measuring.}
\label{tab:param_table}
\end{table}
\section{Conclusion}

In this work, we introduce a powerful fine-matching module based on lightweight yet effective homography estimation. 
By aligning patch pairs from coarse matches, our approach reduces the influence of irrelevant areas, eliminating spatial variance for more precise sub-pixel level matching. 
Furthermore, the ability to freely select matching points on the source patch through homography estimation allows us to maintain keypoint consistency in various tasks and even densify the matching results.
The densification of matching results enables dense supervision during model training, which in turn significantly enhances model performance. 
The optimized version with 5/1/1 parameters outperforms previous methods while also offering greater efficiency.


%

\bigskip
\bibliography{HomoMatcher}

\begin{thebibliography}{37}
\providecommand{\natexlab}[1]{#1}

\bibitem[{Abdel-Aziz, Karara, and Hauck(2015)}]{abdel2015direct}
Abdel-Aziz, Y.~I.; Karara, H.~M.; and Hauck, M. 2015.
\newblock Direct linear transformation from comparator coordinates into object space coordinates in close-range photogrammetry.
\newblock \emph{Photogrammetric engineering \& remote sensing}, 81(2): 103--107.

\bibitem[{Balntas et~al.(2017)Balntas, Lenc, Vedaldi, and Mikolajczyk}]{balntas2017hpatches}
Balntas, V.; Lenc, K.; Vedaldi, A.; and Mikolajczyk, K. 2017.
\newblock HPatches: A benchmark and evaluation of handcrafted and learned local descriptors.
\newblock In \emph{Proceedings of the IEEE conference on computer vision and pattern recognition}, 5173--5182.

\bibitem[{Cao et~al.(2022)Cao, Hu, Sheng, and Shen}]{IHN}
Cao, S.-Y.; Hu, J.; Sheng, Z.; and Shen, H.-L. 2022.
\newblock Iterative deep homography estimation.
\newblock In \emph{Proceedings of the IEEE/CVF Conference on Computer Vision and Pattern Recognition}, 1879--1888.

\bibitem[{Chen et~al.(2024)Chen, Luo, Tian, Bai, Wang, Zhou, Zhen, Fang, Mckinnon, Tsin et~al.}]{chen2024affine}
Chen, H.; Luo, Z.; Tian, Y.; Bai, X.; Wang, Z.; Zhou, L.; Zhen, M.; Fang, T.; Mckinnon, D.; Tsin, Y.; et~al. 2024.
\newblock Affine-based Deformable Attention and Selective Fusion for Semi-dense Matching.
\newblock In \emph{Proceedings of the IEEE/CVF Conference on Computer Vision and Pattern Recognition}, 4254--4263.

\bibitem[{Chen et~al.(2022)Chen, Luo, Zhou, Tian, Zhen, Fang, Mckinnon, Tsin, and Quan}]{chen2022aspanformer}
Chen, H.; Luo, Z.; Zhou, L.; Tian, Y.; Zhen, M.; Fang, T.; Mckinnon, D.; Tsin, Y.; and Quan, L. 2022.
\newblock Aspanformer: Detector-free image matching with adaptive span transformer.
\newblock In \emph{European Conference on Computer Vision}, 20--36. Springer.

\bibitem[{Dai et~al.(2017)Dai, Chang, Savva, Halber, Funkhouser, and Nie{\ss}ner}]{dai2017scannet}
Dai, A.; Chang, A.~X.; Savva, M.; Halber, M.; Funkhouser, T.; and Nie{\ss}ner, M. 2017.
\newblock Scannet: Richly-annotated 3d reconstructions of indoor scenes.
\newblock In \emph{Proceedings of the IEEE conference on computer vision and pattern recognition}, 5828--5839.

\bibitem[{DeTone, Malisiewicz, and Rabinovich(2016)}]{detone2016deep}
DeTone, D.; Malisiewicz, T.; and Rabinovich, A. 2016.
\newblock Deep image homography estimation.
\newblock \emph{arXiv preprint arXiv:1606.03798}.

\bibitem[{DeTone, Malisiewicz, and Rabinovich(2018)}]{detone2018superpoint}
DeTone, D.; Malisiewicz, T.; and Rabinovich, A. 2018.
\newblock Superpoint: Self-supervised interest point detection and description.
\newblock In \emph{Proceedings of the IEEE conference on computer vision and pattern recognition workshops}, 224--236.

\bibitem[{Edstedt et~al.(2023)Edstedt, Athanasiadis, Wadenb{\"a}ck, and Felsberg}]{edstedt2023dkm}
Edstedt, J.; Athanasiadis, I.; Wadenb{\"a}ck, M.; and Felsberg, M. 2023.
\newblock DKM: Dense kernelized feature matching for geometry estimation.
\newblock In \emph{Proceedings of the IEEE/CVF Conference on Computer Vision and Pattern Recognition}, 17765--17775.

\bibitem[{Edstedt et~al.(2024)Edstedt, Sun, B{\"o}kman, Wadenb{\"a}ck, and Felsberg}]{edstedt2024roma}
Edstedt, J.; Sun, Q.; B{\"o}kman, G.; Wadenb{\"a}ck, M.; and Felsberg, M. 2024.
\newblock RoMa: Robust dense feature matching.
\newblock In \emph{Proceedings of the IEEE/CVF Conference on Computer Vision and Pattern Recognition}, 19790--19800.

\bibitem[{Engel, Sch{\"o}ps, and Cremers(2014)}]{engel2014lsd}
Engel, J.; Sch{\"o}ps, T.; and Cremers, D. 2014.
\newblock LSD-SLAM: Large-scale direct monocular SLAM.
\newblock In \emph{European conference on computer vision}, 834--849. Springer.

\bibitem[{Hartley and Zisserman(2003)}]{hartley2003multiple}
Hartley, R.; and Zisserman, A. 2003.
\newblock \emph{Multiple view geometry in computer vision}.
\newblock Cambridge university press.

\bibitem[{He et~al.(2024)He, Sun, Wang, Peng, Huang, Bao, and Zhou}]{he2024detector}
He, X.; Sun, J.; Wang, Y.; Peng, S.; Huang, Q.; Bao, H.; and Zhou, X. 2024.
\newblock Detector-free structure from motion.
\newblock In \emph{Proceedings of the IEEE/CVF Conference on Computer Vision and Pattern Recognition}, 21594--21603.

\bibitem[{Hong et~al.(2022)Hong, Lu, Ye, Lin, Zhao, and Liu}]{hong2022unsupervised}
Hong, M.; Lu, Y.; Ye, N.; Lin, C.; Zhao, Q.; and Liu, S. 2022.
\newblock Unsupervised homography estimation with coplanarity-aware gan.
\newblock In \emph{Proceedings of the IEEE/CVF conference on computer vision and pattern recognition}, 17663--17672.

\bibitem[{Li et~al.(2020)Li, Han, Li, and Prisacariu}]{li2020dual}
Li, X.; Han, K.; Li, S.; and Prisacariu, V. 2020.
\newblock Dual-resolution correspondence networks.
\newblock \emph{Advances in Neural Information Processing Systems}, 33: 17346--17357.

\bibitem[{Li and Snavely(2018)}]{li2018megadepth}
Li, Z.; and Snavely, N. 2018.
\newblock Megadepth: Learning single-view depth prediction from internet photos.
\newblock In \emph{Proceedings of the IEEE conference on computer vision and pattern recognition}, 2041--2050.

\bibitem[{Lowe(2004)}]{lowe2004distinctive}
Lowe, D.~G. 2004.
\newblock Distinctive image features from scale-invariant keypoints.
\newblock \emph{International journal of computer vision}, 60: 91--110.

\bibitem[{Lucas and Kanade(1981)}]{lucas1981iterative}
Lucas, B.~D.; and Kanade, T. 1981.
\newblock An iterative image registration technique with an application to stereo vision.
\newblock In \emph{IJCAI'81: 7th international joint conference on Artificial intelligence}, volume~2, 674--679.

\bibitem[{Mur-Artal, Montiel, and Tardos(2015)}]{mur2015orb}
Mur-Artal, R.; Montiel, J. M.~M.; and Tardos, J.~D. 2015.
\newblock ORB-SLAM: a versatile and accurate monocular SLAM system.
\newblock \emph{IEEE transactions on robotics}, 31(5): 1147--1163.

\bibitem[{Mur-Artal and Tard{\'o}s(2017)}]{mur2017orb}
Mur-Artal, R.; and Tard{\'o}s, J.~D. 2017.
\newblock Orb-slam2: An open-source slam system for monocular, stereo, and rgb-d cameras.
\newblock \emph{IEEE transactions on robotics}, 33(5): 1255--1262.

\bibitem[{Nguyen et~al.(2018)Nguyen, Chen, Shivakumar, Taylor, and Kumar}]{nguyen2018unsupervised}
Nguyen, T.; Chen, S.~W.; Shivakumar, S.~S.; Taylor, C.~J.; and Kumar, V. 2018.
\newblock Unsupervised deep homography: A fast and robust homography estimation model.
\newblock \emph{IEEE Robotics and Automation Letters}, 3(3): 2346--2353.

\bibitem[{Peng et~al.(2022)Peng, Xiang, Fan, Zhao, and Zhao}]{peng2022rwt}
Peng, Q.; Xiang, Z.; Fan, Y.; Zhao, T.; and Zhao, X. 2022.
\newblock RWT-SLAM: Robust visual SLAM for highly weak-textured environments.
\newblock \emph{arXiv preprint arXiv:2207.03539}.

\bibitem[{Rocco, Arandjelovi{\'c}, and Sivic(2020)}]{rocco2020efficient}
Rocco, I.; Arandjelovi{\'c}, R.; and Sivic, J. 2020.
\newblock Efficient neighbourhood consensus networks via submanifold sparse convolutions.
\newblock In \emph{Computer Vision--ECCV 2020: 16th European Conference, Glasgow, UK, August 23--28, 2020, Proceedings, Part IX 16}, 605--621. Springer.

\bibitem[{Rocco et~al.(2018)Rocco, Cimpoi, Arandjelovi{\'c}, Torii, Pajdla, and Sivic}]{rocco2018neighbourhood}
Rocco, I.; Cimpoi, M.; Arandjelovi{\'c}, R.; Torii, A.; Pajdla, T.; and Sivic, J. 2018.
\newblock Neighbourhood consensus networks.
\newblock \emph{Advances in neural information processing systems}, 31.

\bibitem[{Rublee et~al.(2011)Rublee, Rabaud, Konolige, and Bradski}]{rublee2011orb}
Rublee, E.; Rabaud, V.; Konolige, K.; and Bradski, G. 2011.
\newblock ORB: An efficient alternative to SIFT or SURF.
\newblock In \emph{2011 International conference on computer vision}, 2564--2571. Ieee.

\bibitem[{Sarlin et~al.(2019)Sarlin, Cadena, Siegwart, and Dymczyk}]{sarlin2019coarse}
Sarlin, P.-E.; Cadena, C.; Siegwart, R.; and Dymczyk, M. 2019.
\newblock From coarse to fine: Robust hierarchical localization at large scale.
\newblock In \emph{Proceedings of the IEEE/CVF conference on computer vision and pattern recognition}, 12716--12725.

\bibitem[{Sarlin et~al.(2020)Sarlin, DeTone, Malisiewicz, and Rabinovich}]{sarlin2020superglue}
Sarlin, P.-E.; DeTone, D.; Malisiewicz, T.; and Rabinovich, A. 2020.
\newblock Superglue: Learning feature matching with graph neural networks.
\newblock In \emph{Proceedings of the IEEE/CVF conference on computer vision and pattern recognition}, 4938--4947.

\bibitem[{Schonberger and Frahm(2016)}]{schonberger2016structure}
Schonberger, J.~L.; and Frahm, J.-M. 2016.
\newblock Structure-from-motion revisited.
\newblock In \emph{Proceedings of the IEEE conference on computer vision and pattern recognition}, 4104--4113.

\bibitem[{Sun et~al.(2021)Sun, Shen, Wang, Bao, and Zhou}]{sun2021loftr}
Sun, J.; Shen, Z.; Wang, Y.; Bao, H.; and Zhou, X. 2021.
\newblock LoFTR: Detector-free local feature matching with transformers.
\newblock In \emph{Proceedings of the IEEE/CVF conference on computer vision and pattern recognition}, 8922--8931.

\bibitem[{Vaswani(2017)}]{vaswani2017attention}
Vaswani, A. 2017.
\newblock Attention is all you need.
\newblock \emph{arXiv preprint arXiv:1706.03762}.

\bibitem[{Wang et~al.(2024{\natexlab{a}})Wang, Xu, Cui, Wan, and Zhang}]{wang2024fine}
Wang, X.; Xu, R.; Cui, Z.; Wan, Z.; and Zhang, Y. 2024{\natexlab{a}}.
\newblock Fine-grained cross-view geo-localization using a correlation-aware homography estimator.
\newblock \emph{Advances in Neural Information Processing Systems}, 36.

\bibitem[{Wang et~al.(2024{\natexlab{b}})Wang, He, Peng, Tan, and Zhou}]{wang2024efficient}
Wang, Y.; He, X.; Peng, S.; Tan, D.; and Zhou, X. 2024{\natexlab{b}}.
\newblock Efficient LoFTR: Semi-dense local feature matching with sparse-like speed.
\newblock In \emph{Proceedings of the IEEE/CVF Conference on Computer Vision and Pattern Recognition}, 21666--21675.

\bibitem[{Yu et~al.(2023)Yu, Chang, He, Zhang, Yu, and Wu}]{yu2023adaptive}
Yu, J.; Chang, J.; He, J.; Zhang, T.; Yu, J.; and Wu, F. 2023.
\newblock Adaptive spot-guided transformer for consistent local feature matching.
\newblock In \emph{Proceedings of the IEEE/CVF Conference on Computer Vision and Pattern Recognition}, 21898--21908.

\bibitem[{Zaragoza et~al.(2013{\natexlab{a}})Zaragoza, Chin, Brown, and Suter}]{zaragoza2013projective}
Zaragoza, J.; Chin, T.-J.; Brown, M.~S.; and Suter, D. 2013{\natexlab{a}}.
\newblock As-projective-as-possible image stitching with moving DLT.
\newblock In \emph{Proceedings of the IEEE conference on computer vision and pattern recognition}, 2339--2346.

\bibitem[{Zaragoza et~al.(2013{\natexlab{b}})Zaragoza, Chin, Brown, and Suter}]{zaragoza2013asap}
Zaragoza, J.; Chin, T.-J.; Brown, M.~S.; and Suter, D. 2013{\natexlab{b}}.
\newblock As-projective-as-possible image stitching with moving DLT.
\newblock In \emph{Proceedings of the IEEE conference on computer vision and pattern recognition}, 2339--2346.

\bibitem[{Zhao, Huang, and Zhang(2021)}]{zhao2021deep}
Zhao, Y.; Huang, X.; and Zhang, Z. 2021.
\newblock Deep lucas-kanade homography for multimodal image alignment.
\newblock In \emph{Proceedings of the IEEE/CVF conference on computer vision and pattern recognition}, 15950--15959.

\bibitem[{Zhu and Liu(2023)}]{zhu2023pmatch}
Zhu, S.; and Liu, X. 2023.
\newblock Pmatch: Paired masked image modeling for dense geometric matching.
\newblock In \emph{Proceedings of the IEEE/CVF Conference on Computer Vision and Pattern Recognition}, 21909--21918.

\end{thebibliography}



\end{document}